\documentclass[runningheads]{llncs}
\usepackage{graphicx}
\usepackage{comment}
\usepackage{amsmath,amssymb} 
\usepackage{color}

\usepackage[width=122mm,left=12mm,paperwidth=146mm,height=193mm,top=12mm,paperheight=217mm]{geometry}

\usepackage{amssymb}
\usepackage{multirow}
\usepackage{makecell}
\usepackage{amsmath,bm}
\usepackage{rotating}

\begin{document}
\pagestyle{headings}
\mainmatter
\def\ECCVSubNumber{7442}  

\title{Learning Posterior and Prior for Uncertainty Modeling in Person Re-Identification} 

\titlerunning{ECCV-20 submission ID \ECCVSubNumber} 
\authorrunning{ECCV-20 submission ID \ECCVSubNumber} 
\author{Yan Zhang, Zhilin Zheng, Binyu He, Li Sun}
\institute{East China Normal University}

\maketitle

\begin{abstract}
Data uncertainty in practical person reid is ubiquitous, hence it requires not only learning the discriminative features, but also modeling the uncertainty based on the input. This paper proposes to learn the sample posterior and the class prior distribution in the latent space, so that not only representative features but also the uncertainty  
can be built by the model.  
The prior reflects the distribution of all data in the same class, and it is the trainable model parameters. While the posterior is the probability density of a single sample, so it is actually the feature defined on the input.
We assume that both of them are in Gaussian form. To simultaneously model them, we put forward a distribution loss, which measures the KL divergence from the posterior to the priors in the manner of supervised learning. In addition, we assume that the posterior variance, which is essentially the uncertainty, is supposed to have the second-order characteristic. Therefore, a $\Sigma-$net is proposed to compute it by the high order representation from its input. Extensive experiments have been carried out on Market1501, DukeMTMC, MARS and noisy dataset as well.
\end{abstract}

\section{Introduction}
\label{sec:intro}
Person re-identification (reid) is a classic pedestrian retrieval task that aims to find a particular person across non-overlapping camera views \cite{ide}.  Given a query person, reid algorithms find out whether the same person has appeared in the gallery, which contains a large amount of candidates who have emerged at some time, in another place or under a different camera. Usually, the query can be an image \cite{market1501,duke}, a short video sequence \cite{video_based_reid1,mars} and even a text description \cite{text_based_reid1,text_based_reid2}. As reid has a wide application in the surveillance and security system, it has arisen researchers’ attention for years. In the early days, handcrafted features are mainly employed \cite{hand_craft_1,hand_craft_2,hand_craft_3,hand_craft_4}, while nowadays the deep learning based methods \cite{ide,strong_baseline,AOS,ADTA,BFE,CAMA,MLFN} dominate in this area. \cite{ide} is the simplest baseline model using deep features learned from the ResNet50 backbone. Many works \cite{AOS,ADTA,BFE,CAMA,MLFN} extend it and use it for fair comparison. 
Recently, \cite{strong_baseline} summarizes the commonly used tricks for training, and releases another strong baseline model. 

In spite of the tremendous works with different loss functions or network structures, reid is still challenging, due to the low resolutions \cite{low_resolution1,low_resolution2}, various poses \cite{pose_variation1,pose_variation2} or occlusions \cite{AOS}, etc. These difficulties increase the uncertainty for the identification results and 
harm the performance to a certain extent. Most of previous works concentrate on learning deterministic discriminative features for reid \cite{BFE,CAMA,PCB}, but are lack of uncertainty modeling from the data itself. Some works model the data distribution to account the uncertainty \cite{LGM,D,PFE}, while they do not explicitly distinguish the class priors and the sample posteriors. 

This paper proposes to learn the sample posterior and the class priors simultaneously in reid task, so that it quantifies the uncertainty of an input image and its corresponding class. The key idea is to model the feature representation and its uncertainty by two types of probability distribution. The prior models the latent distributions for all samples in the given class. In practice, it is often treated as a set of static parameters. However, our work optimizes them together with other model parameters by back propagation. Moreover, we assume the single Gaussian prior for each class in this paper, which makes the whole setting easily to be expressed in closed form equation. The posterior is the distribution for a single input sample given by the model. It is a special feature using not only for the classification, but also for the uncertainty evaluation. 

Pariticularly, we measure the distance between the posterior and prior by Kullback-Leibler (KL) divergence, which 
serves as the upper bound of the negative log likelihood of the random code, drawn from the posterior, with respect to the given prior. Inspired by GM loss \cite{LGM}, we formulate the class conditional probability by normalizing the negative KL divergence between the posterior and the prior indicated by the label, and  
compute the cross entropy loss. 
An extra regularization term is added to ensure 
the posterior to be closer to the corresponding class prior than others. Furthermore, we propose a structure named $\Sigma-$net to use a high order feature of its input as the variance, so that the mean and variance of the posterior can be distinguishable. 
We perform extensive experiments on the image based reid dataset Market-1501 \cite{market1501} and DukeMTMC-reID \cite{duke} and the video dataset MARS \cite{mars}, and demonstrate the effectiveness of the proposed method.

Our contributions are summarized as follows:
\begin{itemize}
	\item We propose to learn the posterior and the prior to model the uncertainty based on the individual sample and all samples in the same class, respectively.
	\item We propose a novel loss function, named distribution loss, to build the connection between the posterior and the prior distribution.
	\item We design the $\Sigma$-net structure to model the sample uncertainty by computing the high order feature as the variance in the posterior .
	\item We show the effectiveness of our method on several well-known image-based datasets and a video-based dataset.
\end{itemize}

\section{Related Works}
\textbf{High order discriminative feature learning.} To learn discriminative features, many works realize the value of high order features. Non-local attention is one way to exploit the second-order feature, which is also known as the attention mechanism \cite{wang2018non}. \cite{second_order_reid} proposes an inserted non-local attention module to utilize the second order information in the median layers. \cite{ABD_net} proposes a spatial and channel attention from two different directions in the similar way. \cite{auto_reid} 
automatically searches the network structure, and the final results show that the non-local attention module is important for improving the performance. 

Apart from the non-local operation, bilinear pooling \cite{bilinear_pooling_initial,second_order_pooling} is another choice to formulate the high order features. Original bilinear pooling methods have large computation costs. Given the input tensor $\bm{F}\in \mathbb{R}^{H\times W\times C}$, it computes the outer product $\bm{f}^{\text{T}}\bm{f}$, in which $\bm{f}=\bm{F}(i,j)\in \mathbb{R}^{C}$, at each spatial coordinate $(i,j)$. Then a sum pooling is followed to get the second-order feature $\bm{f}^{(2)}=\sum_{i,j}\bm{f}^{\text{T}}\bm{f}\in \mathbb{R}^{C^2}$.  \cite{low_rank_bilnear_pooling} proposes to use Hadamard product to implement it and thus increases the efficiency. In this work, $\bm{F}$ is first reduced to two "thin" tensors $\bm{F}_1$ and $\bm{F}_2$ with only a few channels. Then elementwise product is computed between them to form the second order feature. Similar workes like \cite{horde,mixed_high_order_reid} also proves the effectiveness of Hadamard product. All the above mentioned works validate the high order features, but they are not designed for uncertainty modeling. 

\textbf{Uncertainty modeling by posterior and prior.} In real reid application, there are inevitable noises on the data. These noises are either on the pixels (\emph{e.g.} occlusions, blurs) or on the training labels (\emph{e.g.} ID switch in the tracker). Modeling the data uncertainty mainly intends to deal with these noises. 
Since it improves the robustness of models greatly, it is an important task for its own sake, 
not only for reid. To deal with noisy labels, one type of works \cite{goldberger2016training,northcutt2019confident} aim to find the relation between the noisy and clean labels, and then give the estimated clean labels for training a robust model.  
Another type of works \cite{D,PFE,LGM} model the data ambiguity through the probability distribution, either the sample posterior or the class prior. Besides noisy labels, these works also handle noisy images.  
The model in \cite{D} outputs an auxiliary variance for each sample to increase model robustness when handling with noisy labels. \cite{PFE} evaluates the distance between two images by considering the image-level noise. \cite{D} and \cite{PFE} both assume each sample to be the probability distribution, a conditional Gaussian posterior, in the latent space specified by the model parameters. 

The work in \cite{LGM} models the prior of the whole data as mixture Gaussians, and it uses the same number of independent Gaussians for each class. By maximizing the data likelihood in the corresponding prior distribution, it proposes a so called Gaussian Mixture (GM) loss, 
which performs better than other losses particularly when facing the adversarial samples. 
In GM loss, the extracted feature $\bm{z}$ is assumed to follow a Gaussian mixture distribution as is expressed in Eq. \ref{eq gm px}. $\bm{\mu}^{(k)}$ and $\bm{\sigma}^{(k)}$ are the mean and the variance of class $k$, and $p(k)$ is the prior probability of class $k$. $K$ is the total class number. 
The class probability distribution $p(y|\bm{z})$ can be expressed as Eq. \ref{eq posterior probability}.
\begin{equation}
\label{eq gm px}
p(\bm{z}) = \Sigma_{k=1}^{K}p(\bm{z}|k)p(k) = \Sigma_{k=1}^{K}\mathcal{N}(\bm{z};\bm{\mu}^{(k)}, \bm{\sigma}^{(k)})p(k)
\end{equation}

\begin{equation}
\label{eq posterior probability}
p(y|\bm{z})=\frac{\mathcal{N}(\bm{z};\bm{\mu}^{(y)},\bm{\sigma}^{(y)})p(y)}{\sum_{k=1}^{K}\mathcal{N}(\bm{z};\bm{\mu}^{(k)},\bm{\sigma}^{(k)})p(k)}
\end{equation}
The above works show the feasibility for modeling posterior and prior distributions to cope with the data uncertainty. But the relation between them are still not considered. 



\textbf{Person re-identification.}
The most common data types for person reid task are image and video. Image-based works mainly focus on mining local cues to improve the performance. \cite{PCB} is one of the well-known works that explicitly tear features into parts to enforce the network to mine the discriminative features of different locations to a great extent. \cite{CAMA,AOS,BFE} also apply this idea into their works but in diverse ways. \cite{CAMA} concentrates on enlarging activated areas in CAM, while \cite{AOS,BFE} feed images and features that have been random erased into a network to get rid of the risk of overfitting. To strengthen local information to tackle the overfitting has become a research trend, but few works pay attention to improve the robustness through modeling uncertainty existing in the data. 

For video-based reid, one of the main issue is how to aggregate features in the temporal domain. \cite{DRSA} proposes to use spatiotemporal attention to aggregate features. \cite{Snippet} proposes a similarity aggregation and co-attentive embedding for video-reid. \cite{ADTA} applies attribute information to disentangle the feature embedding and aggregate the temporal features using attention based on the attribute confidence score. \cite{set_distance} makes a sufficient comparison on the effectiveness of different temporal aggregation methods. It propose to calculate the mean and the variance of the temporal features to model the aggregated feature's probability density distribution. This is quite similar to our intuition but it does not model the class priors, and not to apply a network to produce the posterior variance. 

\section{Proposed Methods}
\label{sec:approach}
\subsection{Overview Framework}
\begin{figure}[ht]
	\centering
	\includegraphics[width=1.0\textwidth]{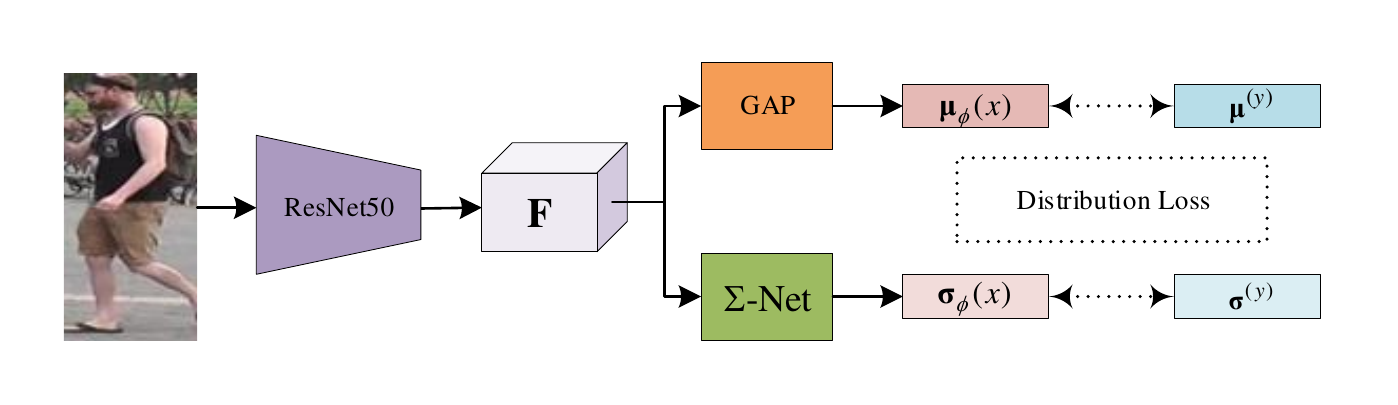}
	\caption{Overview framework of proposed method. It is built upon the backbone of ResNet50. The output $\bm{F}$ is given to two branches. The above one models the mean $\bm{\mu}_{\phi}({\bm{x}})$ of the Gaussian posterior and the bellow one employs the $\Sigma-$net to output the variance $\bm{\sigma}_{\phi}({\bm{x}})$ of the same Gaussian. The posterior Gaussian $q_{\phi}(\bm{z}|\bm{x})=\mathcal{N}(\bm{z};\bm{\mu}_{\phi}(\bm{x}), \bm{\sigma}_{\phi}(\bm{x}))$ is constrained by the prior $p(\bm{z}|y)=\mathcal{N}(\bm{z};\bm{\mu}^{(y)}, \bm{\sigma}^{(y)})$ based on the distribution loss.} 
	\label{fig:framework}
\end{figure}

Fig. \ref{fig:framework} shows the overview framework of our method. The backbone is ResNet50, which is commonly used in reid task \cite{ide}. Its output feature, referred as $\bm{F}\in\mathbb{R}^{H\times W\times C}$, is a 3D tensor depending on the input image $\bm{x}$, with $H$, $W$ and $C$ indicating the height, width and number of channels, respectively. In the upper branch, $\bm{\mu}_{\phi}(\bm{x})\in\mathbb{R}^{C}$ is directly obtained by the global average pooling (GAP), hence it is the mean of $\bm{F}$ in statistics. Note that \cite{ide} directly gives the $\bm{\mu}_{\phi}(\bm{x})$ into a classifier, expecting it to be discriminative, but high order details in $\bm{F}$ are lost. 
Our work exploits $\bm{F}$ in the lower branch, by feeding it into a $\Sigma-$net to compute the high order feature $\bm{\sigma}_{\phi}(\bm{x})$. The idea is to expect $\bm{\sigma}_{\phi}(\bm{x})$ to reflect the uncertainty of the latent code $\bm{z}$ on each corresponding dimension. Here we assume that $\bm{\mu}_{\phi}(\bm{x})$ and $\bm{\sigma}_{\phi}(\bm{x})$ together define the posterior for random variable $\bm{z}\sim q_{\phi}(\bm{z}|\bm{x})$, 
which is of the Gaussian so that $q_{\phi}(\bm{z}|\bm{x})=\mathcal{N}(\bm{z};\bm{\mu}_{\phi}(\bm{x}), \bm{\sigma}_{\phi}(\bm{x}))$. Within this setting, $q_{\phi}(\bm{z}|\bm{x})$ reflects the distribution of $\bm{z}$ evaluated by the model $q_{\phi}$ based on the input image $\bm{x}$. Hence the latent space of $\bm{z}$ becomes probabilistic. 

To complete the classification, we also define the class prior on $\bm{z}$ and assume it to follow the Gaussian form, written as $p(\bm{z}|y)=\mathcal{N}(\bm{z};\bm{\mu}^{(y)}, \bm{\sigma}^{(y)})$. Here the superscript $y$ is the category label of input image $\bm{x}$. Note that all parameters $\bm{\mu}^{(y)}$ and $\bm{\sigma}^{(y)}$ in $p(\bm{z}|y)$, and $\bm{\mu}_{\phi}(\bm{x})$ and $\bm{\sigma}_{\phi}(\bm{x})$ in $q_{\phi}(\bm{z}|\bm{x})$ are of the same dimension, and they are in $\mathbb{R}^C$. 
Different from 
$\bm{\mu}_{\phi}\bm{x})$ and $\bm{\sigma}_{\phi}(\bm{x})$,  $\bm{\mu}^{(y)}$ and $\bm{\sigma}^{(y)}$ are trainable parameters, rather than the output of a certain layer. 
Hence they are expected to evaluate the image posterior $q_{\phi}(\bm{z}|\bm{x})$. Then we design a distribution loss to connect $q_{\phi}(\bm{z}|\bm{x})$ and $p(\bm{z}|y)$, making the posterior from $\bm{x}$ relatively close to the corresponding prior indicated by its class label $y$. Details about the $\Sigma-$net and the distribution loss are given in following two subsections.

\subsection{$\Sigma-$Net}
\label{ssec:sigmanet}
We now provide details about the designed structure, named $\Sigma-$net, for modeling the posterior variance $\bm{\sigma}_{\phi}(\bm{x})$. The basic idea is to adopt the high order features to make $\bm{\sigma}_{\phi}(\bm{x})$ different from the first order feature $\bm{\mu}_{\phi}(\bm{x})$, since the $\bm{\sigma}_{\phi}(\bm{x})$ itself is the second order parameter of a distribution. Actually, mining the high order features becomes common in the fine-grained image classification. Many structures seem to fulfill our requirement. However, these works focus on learning discriminative feature, rather than representing the uncertainty. Here we modify the bilinear pooling structure in \cite{low_rank_bilnear_pooling,horde},  
and change it to $\Sigma-$net so that it becomes appropriate for evaluating the variance of a posterior. 

\begin{figure}[ht]
	\centering
	\includegraphics[width=1.0\textwidth]{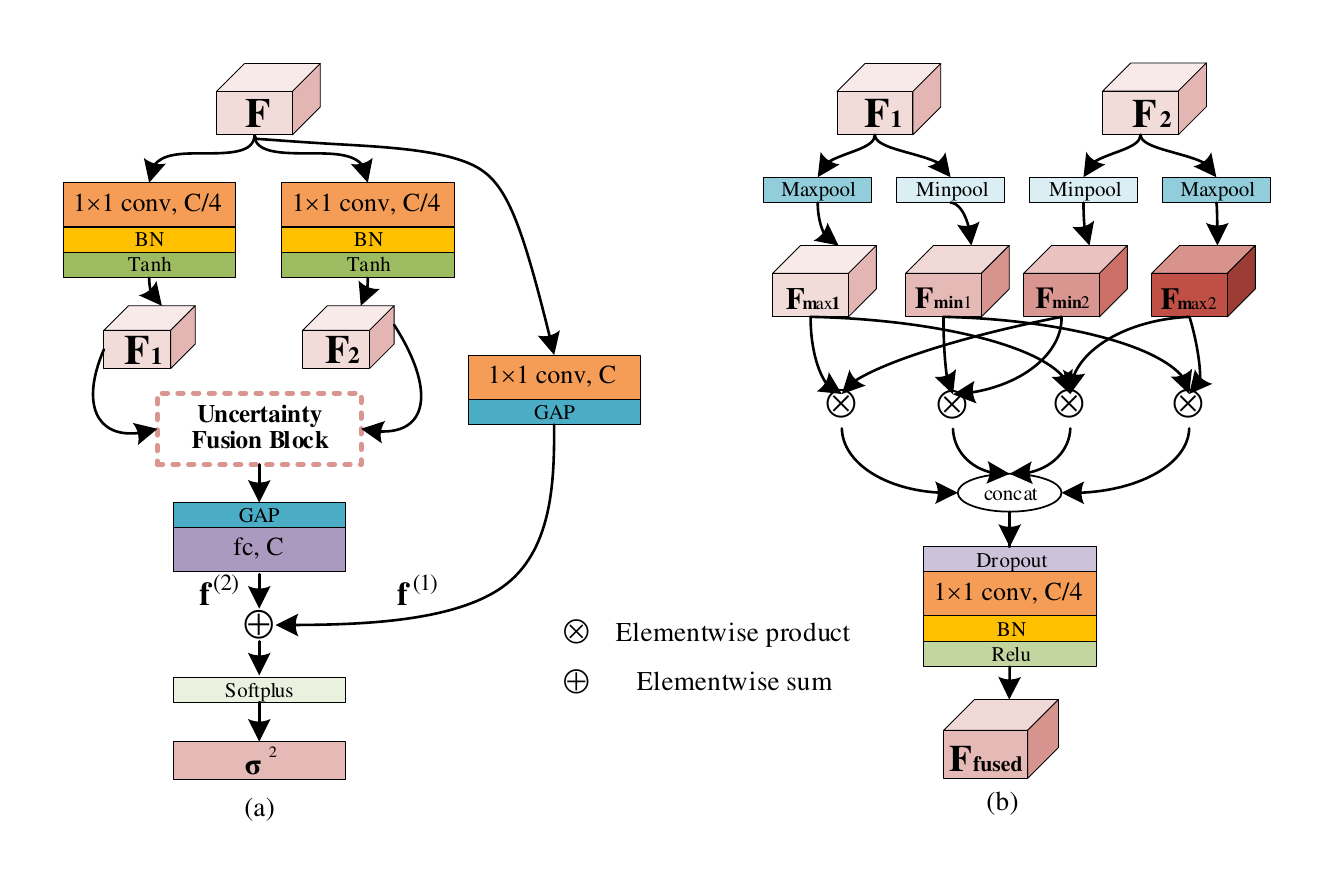}
	\caption{A schematic of the proposed $\Sigma-$net. (a) shows the structure of the whole $\Sigma-$net. It outputs the variance of the posterior based on the second order feature $\bm{f}^{(2)}$ and the first order residual $\bm{f}^{(1)}$. The uncertainty fusion block replaces the direct element-wise production in the blinear pooling layer. (b) illustrates the details of the uncertainty fusion block. It has two inputs $\bm{F}_1$ and $\bm{F}_2$, and gives the output $\bm{F}_{fused}$ with the same size as the two inputs.}
	\label{fig:sigmanet}
\end{figure}

Fig. \ref{fig:sigmanet} shows the $\Sigma-$net which consists of three main branches.
One of them provides the first order residuals $\bm{f}^{(1)}\in\mathbb{R}^C$, while the other two are multiplied together to form the second order features $\bm{f}^{(2)}\in\mathbb{R}^C$. Here we use the superscript to indicate feature order in the current module. Specifically, we first use the $1\times1$ conv to construct a linear shortcut mapping,
and then $\bm{f}^{(1)}$ can be computed simply by average pooling. To form the second order feature, 
instead of directly performing the element-wise multiplication between the feature $\bm{F}_1$ and $\bm{F}_2$ like the work in \cite{low_rank_bilnear_pooling}, we design an uncertainty fusion block which aims to mine the uncertainty in a better way. In this block, the strided min and max pooling on $\bm{F}_1$ and $\bm{F}_2$ is first carried out. So that the local min and max values, within the neighbourhood of a spatial location, are extracted. Hence, two sets of feature maps are obtained in each branch. One is the local min indicated by $\bm{F}_{\min 1}$ and $\bm{F}_{\min 2}$, and the other is the local max $\bm{F}_{\max1}$ and $\bm{F}_{\max2}$. They serve for quantifying the value range at each spatial coordinate. To take advantage of them, we make the cross multiplication between the two branches, that is $\bm{F}_{\min 1}$ (or $\bm{F}_{\max1}$) are multiplied by the $\bm{F}_{\min 2}$ (or $\bm{F}_{\max2}$) in another branch. All together, four groups of feature maps, which include $\bm{F}_{\min 1}\otimes\bm{F}_{\min 2}$, $\bm{F}_{\min 1}\otimes\bm{F}_{\max 2}$, $\bm{F}_{\max 1}\otimes\bm{F}_{\min 2}$, and $\bm{F}_{\max 1}\otimes\bm{F}_{\max 2}$ can be obtained. They are concatenated together followed by the dropout operation, and $1\times 1$ conv-BN-ReLU to reduce the channel number to $C/4$, the same as its input. Finally $\Sigma-$net introduces GAP and another linear function to get the result $\bm{f}^{(2)}\in\mathbb{R}^C$, hence it is of the same dimension with the number of channels of $\bm{F}$. Note that the $\Sigma-$net adopts the softplus function as its activation to make sure that the final second-order feature is positive, and thus suitable for modeling the posterior variance.

\subsection{Distribution Loss}
\label{ssec:distribution_loss}


To build a model reflecting the uncertainty from the data, we assume the network maps its input $\bm{x}$ into a probabilistic embedding defined by the posterior $q_{\phi}(\bm{z}|\bm{x})$, which is in the Gaussian form $\mathcal{N}(\bm{z};\bm{\mu}_{\phi}(\bm{x}), \bm{\sigma}_{\phi}(\bm{x}))$. Here $\phi$ indicates the network parameters that generate the mean and variance of the posterior distribution. 
For the sake of simplicity, we use $\phi$ to represent the parameters related to the posterior, either the mean or the variance. The label $y$ for $\bm{x}$ is also available during training, therefore $q_{\phi}(\bm{z},y|\bm{x})$ is known on training data given $\phi$. On the other hand, we also consider the prior $p(\bm{z}|y)$ (or $p(\bm{z},y)$ given $y$ is known on the training data) on $\bm{z}$ given a specific label $y$. For simplicity, $p(\bm{z}|y)=\mathcal{N}(\bm{z};\bm{\mu}^{(y)}, \bm{\sigma}^{(y)})$ is assumed to be a single Gaussian for each class. Although, both $q_{\phi}(\bm{z}|\bm{x})$ and $p(\bm{z}|y)$ defines the probabilistic characteristic of $\bm{z}$. Their difference is still obvious. The former defines on a particular sample $\bm{x}$, while the latter evaluates whole samples in a class. Usually, the prior $p(\bm{z}|y)$ is predefined and fixed during training. But our work regards $\bm{\mu}^{(y)}$ and $ \bm{\sigma}^{(y)}$ as model parameters indicated by $\psi$, hence they are updated together with other parameters $\phi$ by minimizing the proposed distribution loss. In this case, the prior becomes $p_{\psi}(\bm{z}|y)$ (or $p_{\psi}(\bm{z},y)$).

Note that traditional deterministic embedding is a special case in this assumption. 
If $\bm{\sigma}_{\phi}(\bm{x})\to 0$, $q_{\phi}(\bm{z}|\bm{x})$ becomes a Dirac delta distribution $\delta(\bm{z}-\hat{\bm{z}})$, where $\hat{\bm{z}}$ is the mean of the output feature. This is supposed to happen if the model feels sure about $\bm{x}$. To derive the proposed loss, we need to connect $q_{\phi}(\bm{z}|\bm{x})$ with $p_{\psi}(\bm{z}|y)$ indicated by its label $y$. We choose the KL divergence $D_{KL}(q_{\phi}\|p_{\psi})$ between them as the evaluation metric, 
which turns out to be the negative log likelihood between the extracted feature $\hat{\bm{z}}$ and the assumed prior $p_{\psi}(\bm{z},k)$ when $\bm{\sigma}_{\phi}(\bm{x}) \to 0$. The relation between the likelihood $\mathcal{N}(\hat{\bm{z}};\bm{\mu}^{(k)},\bm{\sigma}^{(k)})$ and the $D_{KL}(q_{\phi}\|p_{\psi})$ when $\bm{\sigma}_{\phi}(\bm{x})\to 0$ is summarized in Eq. \ref{eq likelihood}. The derivation can be found in the Appendix. Note that here $k$ can specify any prior, not necessarily being the same as $y$.
\begin{equation}
\label{eq likelihood}
\mathbb{E}_{q_{\phi}(\bm{z},k|\bm{x})}\mathcal{N}(\bm{z};\bm{\mu}^{(k)},\bm{\sigma}^{(k)})p(k)
=
\exp(-D_{KL}(q_{\phi}(\bm{z},k|\bm{x})\|p_{\psi}(\bm{z}, k))+\text{Const.})
\end{equation}

Combining Eq. \ref{eq likelihood}. with Eq. \ref{eq posterior probability}. from GM loss \cite{LGM} and ignoring the Const., the conditional probability distribution $p(y|\bm{z})$ under our assumption becomes Eq. \ref{eq new posterior probability}. Moreover, the proposed distribution loss can be easily defined by the cross entropy between the $p(y|\bm{z})$ and the one-hot label, as is expressed in Eq. \ref{eq cls}.

\begin{equation}
\label{eq new posterior probability}
p(y|\bm{z})=\frac{\exp(-D_{KL}(q_{\phi}(\bm{z},y|\bm{x})\|p_{\psi}(\bm{z}, y)))}{\sum_{k=1}^{K}\exp(-D_{KL}(q_{\phi}(\bm{z},k|\bm{x})\|p_{\psi}(\bm{z}, k)))}
\end{equation}

\begin{equation}
\label{eq cls}
\begin{split}
&
\mathcal{L}_{cls} = -\frac{1}{N}\sum_{i=1}^{N}\sum_{k=1}^{K}\mathbb{I}(k==y)\log p(k|\bm{z})\\
& 
= -\frac{1}{N}\sum_{i=1}^{N}\log \frac{\exp(-D_{KL}(q_{\phi}(\bm{z},y|\bm{x}_i)\|p_{\psi}(\bm{z}, y)))}{\sum_{k=1}^{K}\exp(-D_{KL}(q_{\phi}(\bm{z},k|\bm{x}_i)\|p_{\psi}(\bm{z}, k)))}
\end{split}
\end{equation}

$\mathcal{L}_{cls} $ only ensures the posterior is relatively closer to the correct prior than the other priors. Similar with GM loss, a KL divergence regularization can be employed to measure to what extent the posterior fits the assumed prior, 
as is shown in Eq. \ref{eq likelihood term}. 
\begin{equation}
\label{eq likelihood term}
\mathcal{L}_{KL} 
=D_{KL}(q_{\phi}(\bm{z},y|\bm{x})\|p_{\psi}(\bm{z}, y))
\end{equation}

In summary, the distribution loss ${L}_{dist}$ can be defined in Eq. \ref{eq distribution loss}, where $\lambda$ is a non-negative hyper parameter. Note that, there is a closed form of KL divergence between two Gaussians, which can be found in the Appendix.

\begin{equation}
\label{eq distribution loss}
\mathcal{L}_{dist} = \mathcal{L}_{cls} + \lambda\mathcal{L}_{KL}
\end{equation}

\subsection{Prior Guided Soft Labels}
In traditional supervised classification, the label is usually represented by a one-hot vector with the number of entries equal to the total number of the classes. Each element is either 1 or 0, with the former lying at the ground truth classes. However, this setting ignores the fact that the distances between different classes are not the same, while these distances can help the classifier to understand the task. Now the key issue is how to model and compute the distance between any two classes. In practice, since it is difficult to evaluate the distance between two classes, the trick of label smoothing is widely used to prevent the classifier from over-fitting. This trick provides the soft label by cutting a small predefined value from the ground truth, and randomly assigning to other elements in the label vector. Indeed, it shows its advantages comparing with the one-hot label, but the class distances are still not considered.   

Our algorithm tries to build the Gaussian priors for each classes in the latent embedding, which implies that the priors naturally lie in the latent space after training, and the distances among them reflect the class similarities. We argue that these Gaussians can be further exploited, and the distance between two of them can serve to form the soft label. Particularly, we employ the Wasserstein distance $D_w$ to evaluate similarity between two class priors $p_{\psi_1}(\bm{z}|y_1)$ and $p_{\psi_2}(\bm{z}|y_2)$. Note that $D_w$ has the closed form solution given two Gaussians, and it can be computed as Eq. \ref{eq w-dist}
\begin{equation}\label{eq w-dist}
D_w(p_{\psi_1}(\bm{z}|y_1);p_{\psi_2}(\bm{z}|y_2))
=\| \bm{\mu}^{(y_1)}-\bm{\mu}^{(y_2)}\| _{2}^{2}
+
\| \sqrt{\bm{\sigma}^{(y_1)}}-\sqrt{\bm{\sigma}^{(y_2)}}\|_{F}^{2}
\end{equation}

Once we have the similarity matrix, we normalize each row with Eq. \ref{eq soft-label}, hence the soft label depending on the learned prior can be generated. $\tau$ is a temperature hyper-parameter, ranging from 0 to 1, for modulating the smoothness of the generated soft label. A higher hyper-parameter brings a softer label. In our experiments, we fix the $\tau$ as $0.17$, with which we can acquire a soft label with its maximum value close to $0.9$. 

\begin{equation}\label{eq soft-label}
\bm{y}_{soft} = \frac{\exp(-D_w/\tau)}{\sum_{k=1}^K\exp(-D_w/\tau)}
\end{equation}

Note that, the soft label is not suitable for an end-to-end training manner as it is unstable due to the rapidly developing priors. We apply the soft label with a two-stage training method. At the first stage, label smoothing is utilized to ensure the model generalization and the training stability. At the second stage, the flexible soft label takes the place of the fixed smoothed label to help finetune the whole model. The evaluation results can be found in section \ref{ssec:ablation}.

\section{Experiments}
\label{sec:experiments}

\subsection{Datasets and Evaluation metrics}
\label{ssec:datasets}
Both the image-based datasets and a video-based dataset are adopted for evaluation. The chosen image-based datasets include Market-1501 \cite{market1501} and DukeMTMC-reID \cite{duke}. The selected video-based dataset is MARS \cite{mars}.  The Cumulative Matching Characteristic (CMC) \cite{cmc} and mean Average Precision (mAP) \cite{market1501} are the performance metrics. 

\textbf{Market-1501} contains 1501 identities captured by 6 cameras from different viewpoints. 12936 images of 751 identities are for training, 3368 query images and 19732 gallery images of the other 750 identities compose the test set. \textbf{DukeMTMC-reID} consists of 36411 images of 1404 identities, half of the identities are used for training, the 2228 query images and 17661 gallery images of the remained identities are applied for testing. \textbf{MARS} is a common dataset used for video-based person re-identification. It consists of 1261 different pedestrians and 20715 tracklets, and each of the identities is captured by at least 2 cameras. We follow the training/evaluation protocol used in \cite{mars}, which selects 625 identities for training and the remaining for testing.

\subsection{Implementation Details}
\label{ssec:implementation}
We extend our experiments on the  \textit{reid strong baseline} framework \cite{strong_baseline}. The used tricks include \textit{warmup}, \textit{random erasing augmentaion}, \textit{label smoothing} and \textit{last stride=1}. Note that, for image-based datasets, we train our model only with the distribution loss, triplet loss is not added, so as to make fair comparison with \cite{D}. While for video-based dataset, we add the triplet loss to match the related works' experiment settings. The total training epoch for MARS is 400 epochs. The base learning rate is 3.5e-4, and it will decay by 3 at 70, 140, 210, 310 epoch. Besides, the number of sequences per batch is 16 and the sequence length is 4.

\subsection{Visualization of the Probabilistic Embedding and the Analysis}
\label{ssec:ablation}

\begin{figure*}[ht]
	\centering
	\includegraphics[width=1.0\textwidth]{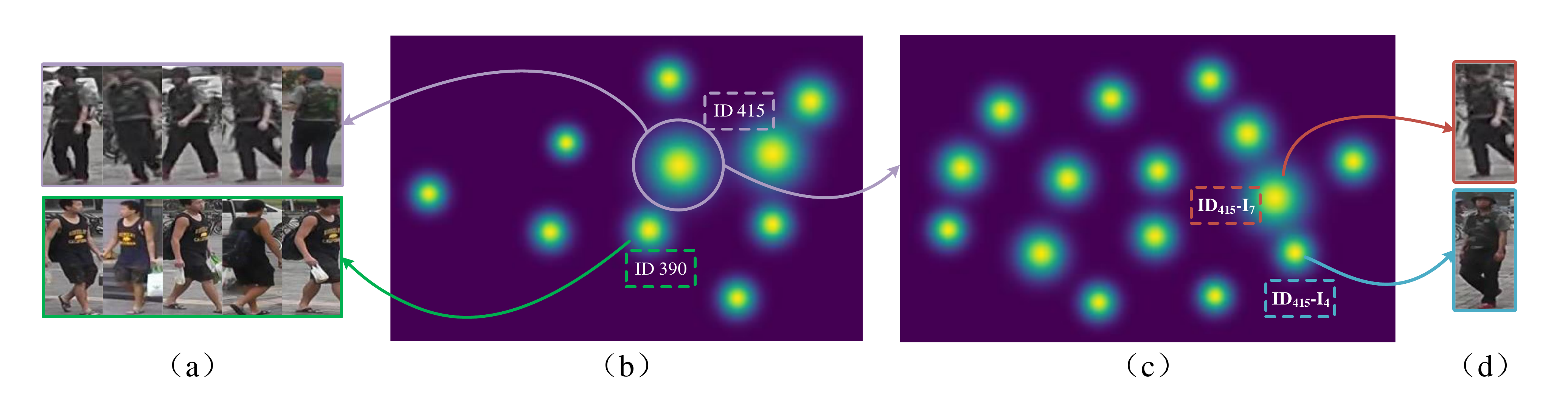}
	\caption{Visualization of the sample posterior and the class prior distribution. (a) The random selected images of ID-415 and ID-390 in Market1501 dataset. (b) The visualization of the prior from 10 randomly selected IDs. (c) The visualization of 15 posteriors in ID-415. (d) Two corresponding raw images of the selected posterior in (c). Radius of the visualized Gaussian distributions is proportional to the corresponding variance, thus larger radius indicates more uncertainty lying in the class or image. This figure is best viewed in color.}
	\label{fig:Visualization}
\end{figure*}

In order to better understand the high dimensional space of the sample posterior and the class prior, we are interested in visualizing them, as is shown in Fig. \ref{fig:Visualization}. Since both the posterior and the prior are represented by high dimensional mean and variance vectors, the visualization is carried out based on the sampling and dimensionality reduction. 
The pipeline is listed as below.
\begin{itemize}
	\item Sample 2000 codes from each distribution.
	\item Use TSNE to project the high-dimensional codes into a 2D plane.
	\item Re-compute the mean and variance of the projected points from a certain  distribution.
	\item Draw a 2D Gaussian distribution with the re-computed mean and variance.
\end{itemize}

In our setting, we model the variance of the prior and the posterior to indicate the uncertainty of the class and the image, respectively. To clearly illustrate the effectiveness of the learned variances of the priors, we randomly choose 10 IDs in Market1501 to visualize the priors for them, which are ID 136, 390, 415, 442, 589, 792, 814, 982, 1007, 1260. Their corresponding priors are shown in Fig. \ref{fig:Visualization} (b). It is obvious that some IDs (\emph{e.g.} ID-415) covering a larger area than the others (\emph{e.g.} ID-390), which implies that it has the larger uncertainty. Note that images from ID-390 are clear and the visual cues, such as the logo on the clothes and the bags in hand and on back, are evident, while images from ID-415 are blurry without many cues, hence it is more ambiguous for identification.

Besides the priors, we are also interested in visualizing the posterior, therefore, we use the same pipelines for visualization on the posteriors from ID-415. The Gaussians for 15 images are shown in Fig.\ref{fig:Visualization} (c). Note that these Gaussians are from images of the same ID-415. But still, we find similar phenomenon as it is on the priors. The covering areas are quite different. 
Fig.\ref{fig:Visualization} (d) shows two raw images ($\text{I}_7$ and $\text{I}_4$) of the corresponding Gaussians in Fig.\ref{fig:Visualization} (c). 
The above image $\text{I}_7$ has the larger variance than the bellow one $\text{I}_4$, and its raw image also has relatively low image quality. It can be concluded that, the learned variances can indicate the uncertainty of a class or an image, and the uncertainty usually depends on the image quality, apparent attributes, etc., which fits well with our common sense.

\subsection{Quantitative Analysis and Ablation Study}
The following experiments can be divided into three aspects. Firstly, we evaluate our propositions on common datasets without noises to see their performance on clean data. Secondly, evaluation on noisy data, consisting of label noise and image noise, is needed to validate the robustness of our model. And finally, we make an ablation study on the designed $\Sigma$-Net.

\subsubsection{Experiments on clean data}
We separately conduct experiments on the baseline model, the class priors and the sample posterior variance, so as to intuitively demonstrate the effectiveness of our proposition. In the following subsections, \textbf{baseline} represents for the classic IDE \cite{ide} model constrained by cross entropy; \textbf{+ prior} means the class priors are added to the baseline and the corresponding loss function is GM loss \cite{LGM}; \textbf{+variance} is our full model, as is shown in Fig. \ref{fig:framework}. \textbf{+ soft label} is considered to be a trick, which can be applied to any model that has defined the class priors. Here, we add the soft label trick to finetune our full model. \textbf{+ reranking} is a widely used trick for person reid, we only add it to the full model finetuned with soft label to see the best performance.


\begin{table}[h]
	\centering
	\caption{The performance of different models evaluated on Market1501 and DukeMTMC-reID datasets}
	\label{performance on imaged-based}
	\begin{tabular}{c|l|cc|cc}
    \hline
    \multicolumn{2}{c|}{\multirow{2}{*}{Methods}} & \multicolumn{2}{c|}{Market1501} & \multicolumn{2}{c}{DukeMTMC-reID} \\
    \multicolumn{2}{c|}{}                         & Rank-1         & mAP            & Rank-1           & mAP             \\ \hline\hline
    \multirow{4}{*}{\rotatebox{90}{G}}        & AOS(CVPR18)\cite{AOS}       & 86.5           & 79.4           & 79.2             & 62.1            \\
                              & DistNet(ICCV2019)\cite{D}       & 87.3           & 70.8           & 74.7             & 56.0            \\
                              & MLFN(CVPR18)\cite{MLFN}      & 90.0           & 74.3           & 81.0             & 62.8            \\
                              & BFE(ICCV19)\cite{BFE}     & \textbf{95.3}  & \textbf{86.7}  & \textbf{89.0}    & \textbf{76.0}   \\ \hline
    \multirow{3}{*}{\rotatebox{90}{L(+G)}}    & HA-CNN(CVPR18)\cite{HA_CNN}    & 91.2           & 75.7           & 80.5             & 63.8            \\
                              & PCB(ECCV2018)\cite{PCB}     & 93.8           & 81.6           & 83.3             & 69.2            \\
                              & CAMA(CVPR19)\cite{CAMA}      & 94.7           & 84.5           & 85.8             & 72.9            \\ \hline\hline
    \multicolumn{2}{c|}{baseline}                 & 89.7           & 78.4           & 80.0             & 65.3            \\
    \multicolumn{2}{c|}{+prior}                   & 89.8           & 78.7           & 80.6             & 68.4            \\
    \multicolumn{2}{c|}{+variance}                & 91.0           & 80.0           & 82.0             & 68.3            \\
    \multicolumn{2}{c|}{+soft label}              & \textbf{91.2}  & \textbf{80.5}  & \textbf{82.6}    & \textbf{68.8}   \\
    \multicolumn{2}{c|}{+reranking}               & 92.4           & 89.8           & 87.3             & 83.7            \\ \hline
    \end{tabular}
	
\end{table}

\begin{table}[h]
	\centering
	\caption{The performance of different models evaluated on MARS dataset}
	\label{performance on MARS}
	\begin{tabular}{l|c|c}
		\hline
		Method          & Rank-1 & mAP  \\ \hline\hline
		DRSA\cite{DRSA}            & 82.3   & 65.8 \\
		Snippet\cite{Snippet}         & 86.3   & 76.1 \\
		ADTA\cite{ADTA}            & \textbf{87.7}   & 78.2 \\ 
		mean\cite{set_distance}              & 82.9   & 76.2 \\
		mean + variance\cite{set_distance}        & 85.2   & 77.9 \\   \hline\hline
		+ prior & 85.4  & 79.3 \\
		+ $\Sigma$-Net    & 85.4   & \textbf{79.6} \\ \hline
	\end{tabular}
	
\end{table}

Our results on both image-based datasets and video-based dataset are listed in Table \ref{performance on imaged-based} and Table \ref{performance on MARS}, respectively. For image-based reid, we divide the previous work into \textbf{G} and \textbf{L (+G)} group. \textbf{G} indicates the work only utilizes the global features, while \textbf{L (+G)} means local features are also employed during training. In our setting, we only use the global features without focusing on the local region. As is shown in Table \ref{performance on imaged-based}, compared with the most of the \textbf{G} methods, our approach has better results, except for BFE \cite{BFE}, which actually applies the drop blocks to enhance the local regions, hence improves the performance of the model. In addition, through separating each proposed modules, it is clearly that each module makes contribution for improving the performance. 

For the video-based work, \cite{set_distance} has proved that modeling sequence feature as a probability density distribution can reach competitive results with other temporal aggregation methods. We then add class priors learning on the GE model \cite{set_distance} by introducing GM loss. As is shown in Table \ref{performance on MARS}, the participation of the class priors brings about an improvement. Furthermore, we exploit the variance of the sample posterior by the proposed $\Sigma$-Net, rather than directly utilize the variance in statistics like \textbf{mean + variance}. The result is listed \textbf{$\Sigma$-Net}, the proposed $\Sigma$-Net brings $0.3\%$ increase in mAP. Before applying the $\Sigma$-Net, we need to first max pool the feature on the temporal space, so that we can get the aggregated feature to produce a variance that is correspond to the input sequence.

\subsubsection{Experiments on noisy data}
\begin{table}[h]
	\centering
	\caption{The performance of different models on Market1501 with $10\%$ random noise on label}
	\label{performance of different models on nosiy dataset}
	\begin{tabular}{l|c|c} 
		\hline
		Method                   & Rank-1        & mAP            \\ 
		\hline\hline
		DistNet\cite{D}         & 82.1          & 62.0           \\
		baseline                 & 79.4          & 58.9           \\
		+prior           & 81.8          & 63.3  \\
		+variance & \textbf{83.1} & \textbf{65.8}           \\
		\hline
	\end{tabular}
	
\end{table}
Data uncertainty usually reflects on two aspects, "noises on image" or "noises on label". DistNet \cite{D} has conducted sufficient experiment on modeling a posterior distribution to deal with noisy data. We follow the testing protocol proposed in \cite{D} to evaluate our approach on Market1501 with $10\%$ random noise. To be fair, we re-implement the DistNet net on this dataset, and the result listed in Table \ref{performance of different models on nosiy dataset} is slightly higher than the results claimed in \cite{D}. When the class priors are added to the baseline, the mAP has already exceed DistNet $1.3\%$. And then by adding the posterior variance, our approach can still have $2\%$ increase in Rank-1 and mAP. The results once again prove the effectiveness of our model.

To evaluate the robustness against the image-level noise, we design a new testing protocol. As person reid is a quite easy to overfit, at the training stage, most works prefer to apply random crop, erasing, flip, etc. to strengthen the generalization of their models. Therefore, at the test stage, the noise from the augmentation will not be a challenge for the models. To avoid the influence from the data augmentation, we choose to add Gaussian blur with different kernel size onto the raw query images. And then the blurred query images will be used for retrieving among the clean gallery images. We have evaluated our model on four different degrees of blur, and the evaluation results are listed in Table \ref{noisy image}. Note that, all the models to be evaluated are trained on the clean data. It can be seen that the learned class priors make few contribution against the blurry attack, and the sample posterior plays a key role for strengthening model robustness. DistNet also models the sample posterior, but poor performance may be due to a certain amount of information loss caused by the sampling operation.

\begin{table}[]
\centering
\caption{The performance of different models on Market1501 with four Gaussian blur of different kernel size}
\label{noisy image}
\begin{tabular}{c|cc|cc|cc|cc}
\hline
\multirow{2}{*}{Method} & \multicolumn{2}{c|}{$0\times0$} & \multicolumn{2}{c|}{$3\times3$} & \multicolumn{2}{c|}{$5\times5$} & \multicolumn{2}{c}{$7\times7$} \\ \cline{2-9} 
                        & Rank-1          & mAP          & Rank-1          & mAP           & Rank-1          & mAP           & Rank-1          & mAP           \\ \hline\hline
DistNet \cite{D}        & 87.6            & 72.8         & 82.4            & 66.9          & 62.4            & 48.1          & 35.7            & 27.8          \\
baseline                & 89.7            & 78.4         & 85.7            & 73.8          & 62.7            & 50.6          & 38.3            & 30.3          \\
+ prior                 & 89.8            & 78.7         & 84.4            & 72.5          & 59.8            & 48.7          & 29.7            & 24.6          \\
+ variance              & \textbf{91.0}            & \textbf{80.0}         & \textbf{86.8}            & \textbf{74.9}          & \textbf{69.6}            & \textbf{56.8}          & \textbf{43.6}            & \textbf{35.1}          \\ \hline
\end{tabular}
\end{table}

\begin{figure*}[ht]
	\centering
	\includegraphics[width=0.8\textwidth]{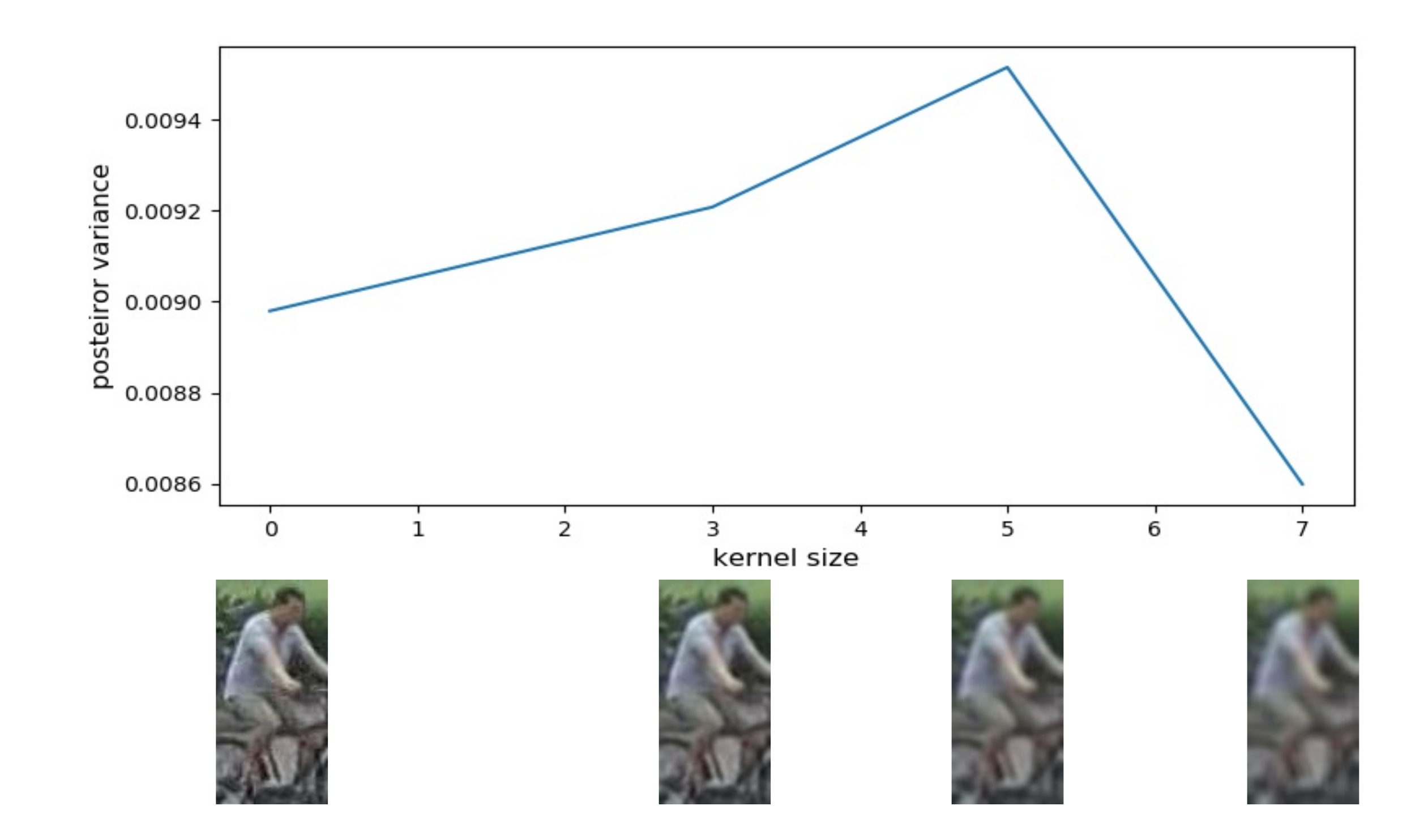}
	\caption{Examples of blurred query images and the mean value of their corresponding posterior variances. From left to the right, the blurred kernel size is $0\times0$ (\emph{i.e.} raw images), $3\times3$, $5\times5$ and $7\times7$. Best viewed in color.}
	\label{fig:blur_sigma}
\end{figure*}

In Fig. \ref{fig:blur_sigma}, we list four blurry examples and the mean value of their corresponding posterior variance on different dimensions. As the degree of ambiguity deepens, the output posterior variance will gradually increase. But after a certain level, it will drop rapidly. We consider it is because the sever ambiguity has successfully attack the model, and the posterior distribution starts to be away from the previous center.

\subsubsection{Ablation study on $\Sigma$-Net.}

\begin{table}[]
\centering
	\label{ablation study on sigmanet}
	\caption{The ablation study on sigma net. The performance is evaluated on Market1501 dataset}
\begin{tabular}{c|cc|cc|cc|cc|ll}
\hline
\multirow{2}{*}{Method}   & \multicolumn{2}{c|}{clean data} & \multicolumn{2}{c|}{noisy label} & \multicolumn{2}{c|}{noisy image (3)} & \multicolumn{2}{c|}{noisy image (5)} & \multicolumn{2}{l}{noisy image (7)} \\ \cline{2-11} 
                          & Rank-1         & mAP           & Rank-1          & mAP            & Rank-1            & mAP              & Rank-1            & mAP              & Rank-1           & mAP              \\ \hline\hline
MLP                       & 90.5           & 80.0          & 82.2            & 61.8           & 84.5              & 70.5             & 62.4              & 49.1             & 35.8             & 27.9             \\
BM            & \textbf{90.8}  & \textbf{80.4} & \textbf{84.2}   & 63.2           & 85.0              & 71.2             & 63.5              & 50.8             & 37.4             & 29.9             \\
$\Sigma$-Net & 91.0           & 80.0          & 83.1            & \textbf{65.8}  & \textbf{86.8}     & \textbf{74.9}    & \textbf{69.6}     & \textbf{56.8}    & \textbf{43.6}    & \textbf{35.1}  \\ \hline
\end{tabular}
\end{table}

The proposed $\Sigma$-Net has gone through three development period. The first one is an \textbf{MLP}, which only generates the first-order feature. Then we consider to use the bilinear model (\textbf{BM}), which refers to the $\Sigma$-Net without the uncertainty fusion block, to introduce the second-order feature. And in the last period, we add the \textbf{Uncertainty Fusion Block} to better grasp the feature uncertainty. Table \ref{ablation study on sigmanet} lists the comparison results on the three structures. When evaluated on clean data, the three structures have the similar performance, and \textbf{BM} seems to be the best. However, when we apply them to noisy data, $\Sigma$-Net has the robustest characteristic. 


\section{Conclusion}
In this paper, we propose to learn the posterior and prior simultaneously with a novel distribution loss, which builds the connection between the posterior and the prior and makes the whole training process able to follow the end-to-end training trend. Furthermore, we propose a $\Sigma$-Net module to output a second-order feature as the posterior variance to maintain the mathematical significance of the variance, which is an omitted point of the previous work. In addition, we visualize the learned posteriors and class priors, illustrating that the learned distributions can reflect the data characteristics to some extent. We also list the results of comparison with other works and ours ablation studies. The experiment results meet our expectation and prove the effectiveness of our approach.

\clearpage

\appendix
\section{Mathematical proofs}
We would like to set up the relation between the proposed distribution loss and the GM loss \cite{LGM}. It is easy to prove that the KL divergence, between the posterior $q_{\phi}(\bm{z}|\bm{x})$ and the prior $p(\bm{z}|y)$, degenerates to the log-likelihood of the prior $p(\bm{z}|y)$, when the posterior variance $\bm{\sigma}_{\phi}(\bm{x})\to 0$. On the other hand, the we assume both $q_{\phi}(\bm{z}|\bm{x})$ and $p(\bm{z}|y)$ are Gaussians, therefore, the KL divergence can be computed in the closed-form solution. These details are given in \ref{A:1} and \ref{A:2}, respectively. 
		\subsection{Relation between KL divergence and log-likelihood}\label{A:1}
		 Assume that the posterior $q_{\phi}(\bm{z}|\bm{x})$ is a Dirac delta distribution.
		$$q_{\phi}(\bm{z}|\bm{x})=\delta(\bm{z}-\hat{\bm{z}})$$
		If we assume conditional independence between $\bm{z}$ and the class $y$, then we have
		$$q_{\phi}(\bm{z},y|\bm{x})=q_{\phi}(\bm{z|\bm{x}})p(y|\bm{x})$$
		$p(y|\bm{x})$ is the one-hot encoding of label.
		
		\begin{equation*}
		\begin{split}
		&D_{KL}(q_{\phi}(\bm{z},y|\bm{x})\|p(\bm{z},y))\\
		&=D_{KL}[\delta(\bm{z}-\hat{\bm{z}})p(y|\bm{x})\|p(\bm{z}|y)p(y)]\\
		&=-\sum_{k}^{}\int\delta(\bm{z}-\hat{\bm{z}})p(k|\bm{x})\log\frac{p(\bm{z}|k)p(k)}{\delta(\bm{z}-\hat{\bm{z}})p(k|\bm{x})}\text{d}z\\
		&=-\sum_{k}^{}\int\delta(\bm{z}-\hat{\bm{z}})p(k|\bm{x})\log p(\bm{z}|k)\text{d}z -\sum_{k}^{}\int\delta(\bm{z}-\hat{\bm{z}})p(k|\bm{x})\log \frac{p(k)}{p(k|\bm{x})}\text{d}z\\
		&+\sum_{k}\int\delta(\bm{z}-\hat{\bm{z}})p(k|\bm{x})\log\delta(\bm{z}-\hat{\bm{z}})\text{d}z\\
		&=-\sum_{k}p(k|\bm{x})\log p(\hat{\bm{z}}|k)-\sum_{k}p(k|\bm{x})\log\frac{p(k)}{p(k|\bm{x})}+\int\delta(\bm{z}-\hat{\bm{z}})\log \delta(\bm{z}-\hat{\bm{z}})\text{d}z
		\end{split}
		\end{equation*}
		
		The last two terms are constants, therefore,
		
		\begin{equation*}
		\begin{split}
	D_{KL}(q_{\phi}(\bm{z},y|\bm{x})\|p(\bm{z},y)))
		&=-\sum_{k}p(k|\bm{x})\log p(\hat{\bm{z}}|k)+\text{Const.}\\
		&=-\sum_{k}\mathbb{I}(k==y)\log p(\hat{\bm{z}}|k)+\text{Const.}\\
		&=-\sum_{k}\mathbb{I}(k==y)\log\mathcal{N}(\hat{\bm{z}};\bm{\mu}^{(k)},\bm{\sigma}^{(k)})+\text{Const.}\\
		&=-\log\mathcal{N}(\hat{\bm{z}};\bm{\mu}^{(y)},\bm{\sigma}^{(y)})+\text{Const.}
		\end{split}
		\end{equation*}
		
		Hence, 
		
		$$\mathcal{N}(\hat{\bm{z}};\bm{\mu}^{(y)},\bm{\sigma}^{(y)})=\exp(-D_{KL}(q_{\phi}(\bm{z},y|\bm{x})\|p(\bm{z},y)+\text{Const.})$$
		
		Note that, this equation is equivalent to Eq. 3 in the main body, as the expectation in Eq. 3 can be omitted due to $p(k)$ is a one-hot label.

		
		
		\subsection{KL divergence of two multivariate Gaussians}\label{A:2}
		
		$$D_{KL}(f\|g)=\int f(x)\ln\frac{f(x)}{g(x)}\text{d}x$$
		
		For two Gaussians $f$ and $g$ defined as the $d$ dimensional probability density functions,  the KL divergence has a closed formed expression, where $\bm{\mu}_g$,  $\bm{\Sigma}_g$ and $\bm{\mu}_f$,  $\bm{\Sigma}_f$ are the mean vectors and co-variance matrices for $g$ and $f$, respectively.
		$$D(f\|g)=\frac{1}{2}[\ln\frac{|\bm{\Sigma}_g|}{|\bm{\Sigma}_f|}
		+Tr(\bm{\Sigma}_{g}^{-1}\bm{\Sigma}_{f})-d
		+(\bm{\mu}_{f}-\bm{\mu}_{g})^{\top}\bm{\Sigma}_{g}^{-1}(\bm{\mu}_{f}-\bm{\mu}_{g})]$$
		
	Hence, in our setting, 
		\begin{equation*}
		\begin{split}
		&D(q_{\phi}(\bm{z}|\bm{x})\|p(\bm{z}|y)\\
		&=\frac{1}{2}[\ln\frac{|\bm{\sigma}^{(y)}|}{|\bm{\sigma}_{\phi}(\bm{x})|}
		+Tr(\frac{\bm{\sigma}_{\phi}(\bm{x})}{\bm{\sigma}^{(y)}})
		+\frac{(\bm{\mu}_{\phi}(\bm{x})-\bm{\mu}^{(y)})^{\top}(\bm{\mu}_{\phi}(\bm{x})-\bm{\mu}^{(y)})}{\bm{\sigma}^{(y)}}+\text{Const.}
		\end{split}
		\end{equation*}
		
	Note that $\bm{\mu}_{\phi}(\bm{x})$, $\bm{\sigma}_{\phi}(\bm{x})$, $\bm{\mu}^{(y)}$ and $\bm{\sigma}^{(y)}$ are diagonal matrices.

\section{Supplementary Experiments}

\begin{figure}[ht]
	\centering
	\includegraphics[width=0.7\textwidth]{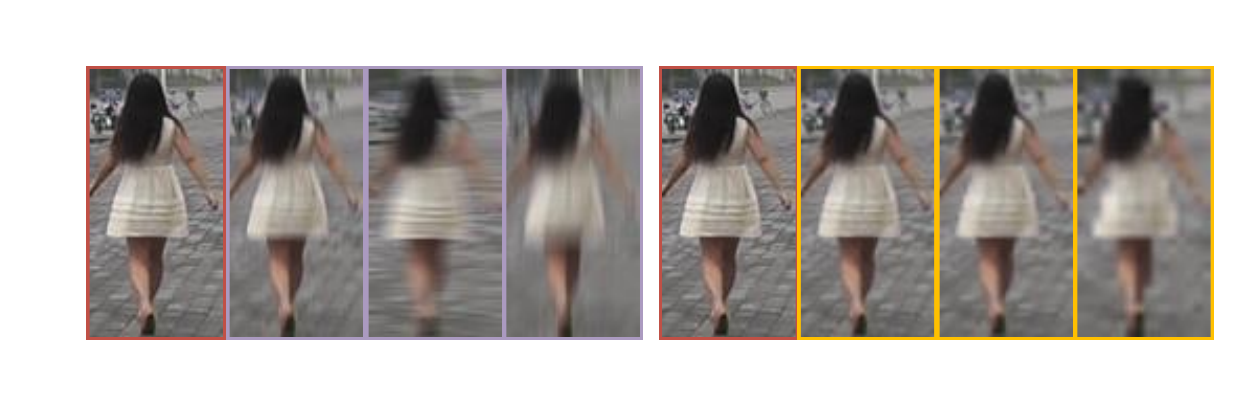}
	\caption{Degraded images from the motion blur (the left four) and the interpolation noise (the right four). Red boxes are the raw images. Purple boxes, from left to the right, correspond to noise levels with $5\times 5$, $10\times 10$, $15\times 15$ kernel size. Similarly, images in yellow boxes, from left to right, are the decayed results of the bi-linear resizing from the downsize at 0.75, 0.5 and 0.25 of the original resolution. Best viewed in color.}
	\label{fig:blur}
\end{figure}

\begin{figure}[ht]
	\centering
	\includegraphics[width=0.7\textwidth]{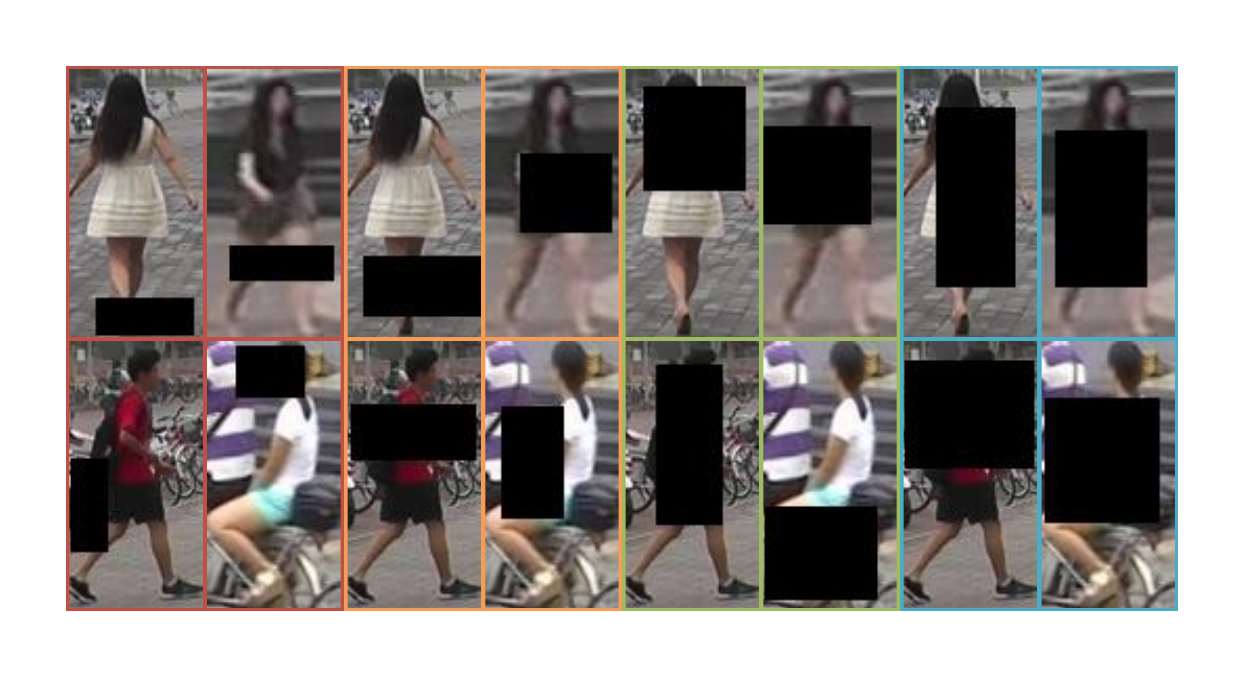}
	\caption{Samples of randomly erased images with different area sizes. From the left to the right, there are four groups of images. Group red contains images with $10\%$ of the randomly erased area. Group orange, green and blue are for $20\%$, $30\%$ and $40\%$, respectively.}
	\label{fig:random erasing}
\end{figure}

To further validate our propositions' robustness against image-level noise, we conduct the following experiments. We manually make up three different types of noises, which are the simulations of motion blurs, low resolutions and occlusions, 
to degrade the image quality of the query set in the Market1501 \cite{market1501}. The degraded images 
are shown in Fig. \ref{fig:blur} and Fig. \ref{fig:random erasing}, respectively. Note that to demonstrate the robustness of model for unseen noises, the degraded images do not incorporate into the training set by the data augmentations.

\begin{table}[]
\centering
\caption{The performances of different models against the motion blur on Market1501.}
\label{motion blur}
\begin{tabular}{c|cc|cc|cc|cc}
\hline
\multirow{2}{*}{Method}           & \multicolumn{2}{c|}{$0\times0$} & \multicolumn{2}{c|}{$5\times5$}                                 & \multicolumn{2}{c|}{$10\times10$}                               & \multicolumn{2}{c}{$15\times15$}                               \\ \cline{2-9} 
                                  & Rank-1         & mAP           & Rank-1                         & mAP                            & Rank-1                         & mAP                            & Rank-1                         & mAP                            \\ \hline
DistNet \cite{D} & 87.6           & 72.8          & 76.8                           & 60.3                           & 46.3                           & 35.5                           & 21.1                           & 17.2                           \\
baseline                          & 89.7           & 78.4          & 79.7                           & 67.2                           & 49.6                           & 40.7                           & 20.0                           & 19.4                           \\
+ prior                           & 89.8           & 78.7          & 79.6                           & 67.1                           & 46.2                           & 37.8                           & 19.6                           & 16.9                           \\
+ variance                        & \textbf{91.0}  & \textbf{80.0} & \textbf{83.0} & \textbf{70.7} & \textbf{52.8} & \textbf{43.4} & \textbf{23.6} & \textbf{21.1} \\ \hline
\end{tabular}
\end{table}

\begin{table}[]
\centering
\caption{The performance of different models against the interpolation noise on Market1501. }
\label{bilinear interpolation noise}
\begin{tabular}{c|cc|cc|cc|cc}
\hline
\multirow{2}{*}{Method} & \multicolumn{2}{c|}{1.0}      & \multicolumn{2}{c|}{0.75}     & \multicolumn{2}{c|}{0.50}     & \multicolumn{2}{c}{0.25}     \\ \cline{2-9} 
                        & Rank-1        & mAP           & Rank-1        & mAP           & Rank-1        & mAP           & Rank-1        & mAP           \\ \hline
DistNet \cite{D}                 & 87.6          & 72.8          & 83.2          & 67.1          & 78.4          & 61.9          & 56.8          & 43.1          \\
baseline                & 89.7          & 78.4          & 87.9          & 75.7          & 83.0          & 69.2          & 56.9          & 45.8          \\
+ prior                 & 89.8          & 78.7          & 87.1          & 75.9          & 81.9          & 69.7          & 54.1          & 43.7          \\
+ variance              & \textbf{91.0} & \textbf{80.0} & \textbf{88.7} & \textbf{77.8} & \textbf{85.2} & \textbf{72.8} & \textbf{62.9} & \textbf{51.5} \\ \hline
\end{tabular}
\end{table}

\begin{figure}[ht]
	\centering
	\includegraphics[width=0.4\textwidth]{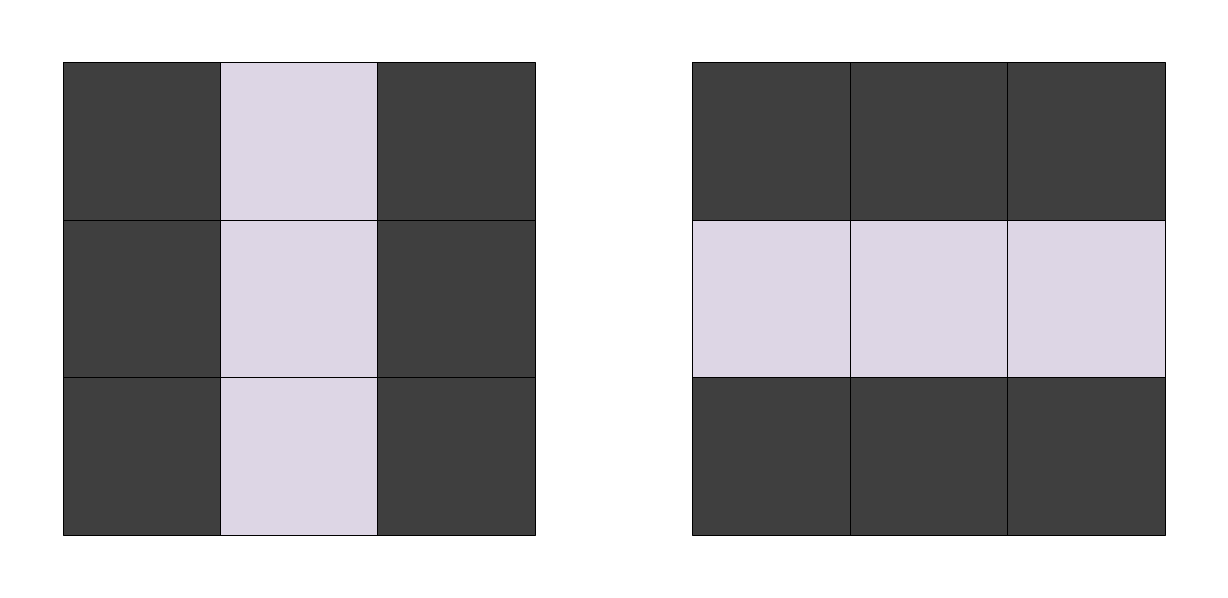}
	\caption{Examples of $3\times 3$ motion blur kernel. The left kernel is for generating vertical motion blur, and the right is for horizontal motion blur. The value of the black element is 0, and the value of the purple one is $1 / kernel size$.}
	\label{fig:blur kernel}
\end{figure}

Motion blur often happens in reid 
due to the targets' fast moving. In Table \ref{motion blur}, the kernel sizes of  $\bm{5\times 5}$, $\bm{10\times 10}$, and $\bm{15\times 15}$ indicate the intensity of the blurring. 
The kernel are essentially of two types, as is shown in Fig. \ref{fig:blur kernel}, 
to simulate the blur caused by the vertical and horizontal motions,  respectively. The probability of applying them are $50\%$ and $50\%$. 

The second type of degradation is caused by the limited resolution of target. 
To simulate this case, we first downsize the original image, then apply the bi-linear interpolation to recover it. 
The test results are listed in Table \ref{bilinear interpolation noise}, \textbf{0.75}, \textbf{0.50}, \textbf{0.25} represent for the different downsize ratios. 

It can be seen from Table \ref{motion blur} and Table \ref{bilinear interpolation noise}, our full model achieves the best performances in both cases. Here \textbf{baseline}, \textbf{+prior} and \textbf{+variance} are the three ablation models, which are introduced in experimental section of the paper. \textbf{DistNet} is the model from \cite{D}. 

\begin{table}[]
\centering
\caption{The performance of different models against the random erasing on Market1501. }
\label{random erasing}
\begin{tabular}{c|cc|cc|cc|cc|cc}
\hline
\multirow{2}{*}{Method} & \multicolumn{2}{c|}{0.}        & \multicolumn{2}{c|}{0.1}      & \multicolumn{2}{c|}{0.2}      & \multicolumn{2}{c|}{0.3}      & \multicolumn{2}{c}{0.4}                              \\ \cline{2-11} 
                        & Rank-1        & mAP           & Rank-1        & mAP           & Rank-1        & mAP           & Rank-1        & mAP           & \multicolumn{1}{c}{Rank-1} & \multicolumn{1}{c}{mAP} \\ \hline
DistNet \cite{D}                & 83.3          & 65.9          & 41.5          & 32.8          & 20.0          & 17.0          & 9.2           & 8.7           & 3.8                        & 3.8                     \\
baseline                & 86.7          & 72.4          & 47.6          & 39.1          & 25.9          & 21.0          & 9.8           & 9.6           & 3.6                        & 4.0                     \\
+ prior                 & \textbf{88.9} & \textbf{74.9} & \textbf{58.7} & \textbf{47.6} & 30.6          & 26.5          & 12.2          & 12.0          & 4.1                        & 4.8                     \\
+ variance              & 87.9          & 73.8          & 55.4          & 45.1          & \textbf{31.2} & \textbf{26.8} & \textbf{14.9} & \textbf{13.9} & \textbf{5.7}               & \textbf{6.4}            \\ \hline
\end{tabular}
\end{table}

The third type of noise we are considering is the occlusion. And most classic works prefer to use random erasing as an augmentation in the training phase. To fairly evaluate the robustness against the occlusion, we retrain all the four compared models without the random erasing augmentation, 
and apply it only on the query set during the evaluation. 
As is shown in Fig. \ref{fig:random erasing}, we control the erasing area as the noise level. 

Table \ref{random erasing} lists the comparison results of random erasing test. When the occluded area is small, the \textbf{+ prior} model shows the best performance. While as it 
becomes larger, our full model shows its potential against the occlusion attack. 

%
%

\bibliographystyle{splncs04}
\bibliography{egbib}
\end{document}